\documentclass{article} 
\usepackage{collas2024_conference,times}
\usepackage{easyReview}

\usepackage{hyperref}
\hypersetup{
    colorlinks=true,
    linkcolor=red,
    filecolor=magenta,
    urlcolor=blue,
    citecolor=purple,
    pdftitle={Overleaf Example},
    pdfpagemode=FullScreen,
    }
\usepackage{amsthm}
\usepackage{amsmath}
\usepackage{amsfonts}
\usepackage{amssymb}
\usepackage{graphicx}
\usepackage{subcaption}
\usepackage{algorithm}
\usepackage{algpseudocode}
\usepackage{booktabs}
\usepackage{float}

\newcommand{\fs}[1]{\footnotesize $\pm$#1}

\title{t-DGR: A Trajectory-Based \\
Deep Generative Replay Method for \\
Continual Learning in Decision Making}

\author{William Yue \\
The University of Texas at Austin \\
\texttt{william.yue@utexas.edu} \\
\And 
Bo Liu  \\
The University of Texas at Austin \\
\texttt{bliu@cs.utexas.edu} \\
\And 
Peter Stone \\
The University of Texas at Austin \\
Sony AI \\
\texttt{pstone@cs.utexas.edu}
}

\collasfinalcopy 

\begin{document}

\maketitle

\begin{abstract}
Deep generative replay has emerged as a promising approach for continual learning in decision-making tasks. This approach addresses the problem of catastrophic forgetting by leveraging the generation of trajectories from previously encountered tasks to augment the current dataset. However, existing deep generative replay methods for continual learning rely on autoregressive models, which suffer from compounding errors in the generated trajectories. In this paper, we propose a simple, scalable, and non-autoregressive method for continual learning in decision-making tasks using a generative model that generates task samples conditioned on the trajectory timestep. We evaluate our method on Continual World benchmarks and find that our approach achieves state-of-the-art performance on the average success rate metric among continual learning methods.
\end{abstract}

\section{Introduction}
Continual learning, also known as lifelong learning, is a critical challenge in the advancement of general artificial intelligence, as it enables models to learn from a continuous stream of data encompassing various tasks, rather than having access to all data at once \citep{Ring:1994}. However, a major challenge in continual learning is the phenomenon of catastrophic forgetting, where previously learned skills are lost when attempting to learn new tasks \citep{MCCLOSKEY1989109}.

To mitigate catastrophic forgetting, replay methods have been proposed, which involve saving data from previous tasks and replaying it to the learner during the learning of future tasks. This approach mimics how humans actively prevent forgetting by reviewing material for tests and replaying memories in dreams. However, storing data from previous tasks requires significant storage space and becomes computationally infeasible as the number of tasks increases.

In the field of cognitive neuroscience, the Complementary Learning Systems theory offers insights into how the human brain manages memory. This theory suggests that the brain employs two complementary learning systems: a fast-learning episodic system and a slow-learning semantic system \citep{mcclelland1995complementary, KUMARAN2016512, cls}. The hippocampus serves as the episodic system, responsible for storing specific memories of unique events, while the neocortex functions as the semantic system, extracting general knowledge from episodic memories and organizing it into abstract representations \citep{4f488fef93a149c3894ef77d95fc1653}.

\begin{figure*}[t!]
\includegraphics[width=\linewidth]{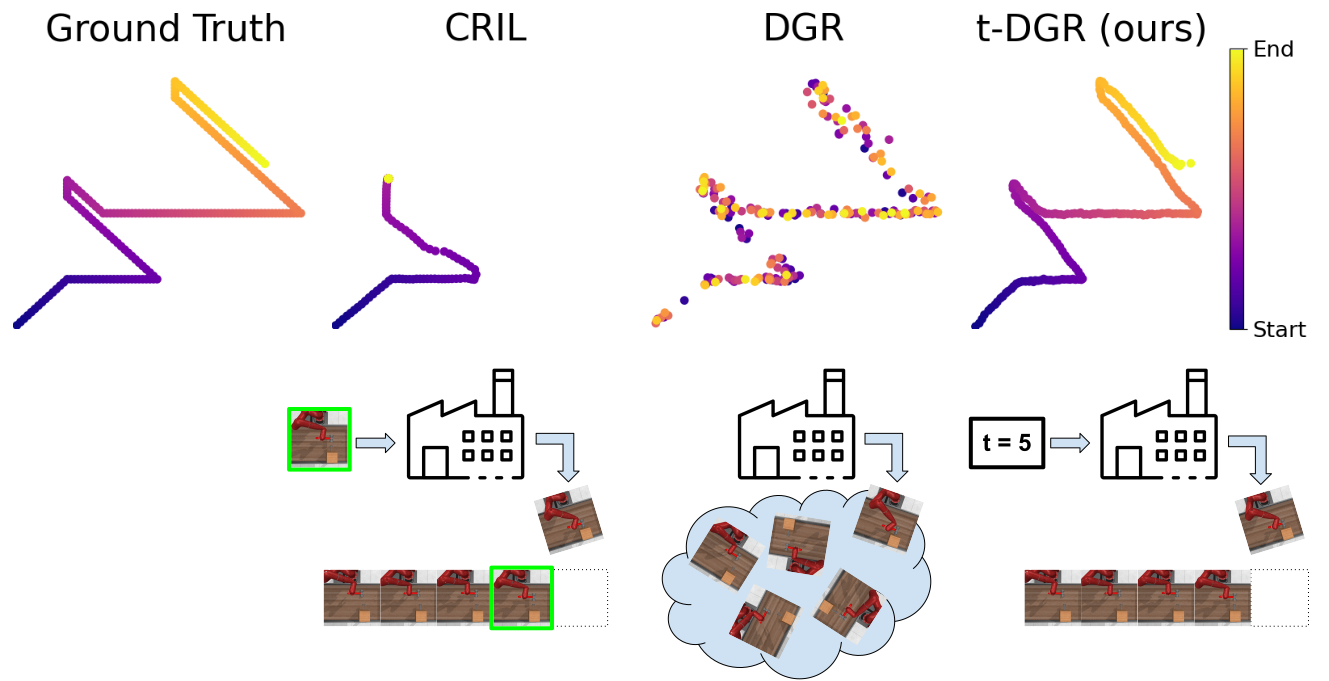}
\caption{The first row presents a comparison of three generative methods for imitating an agent's movement in a continuous 2D plane with Gaussian noise. The objective is to replicate the ground truth path, which transitions from darker to lighter colors. The autoregressive method (CRIL) encounters a challenge at the first sharp turn as nearby points move in opposing directions. Once the autoregressive method deviates off course, it never recovers and compromises the remaining trajectory. In contrast, sampling individual state observations i.i.d. without considering the temporal nature of trajectories (DGR) leads to a fragmented path with numerous gaps. Our proposed method t-DGR samples individual state observations conditioned on the trajectory timestep. By doing so, t-DGR successfully avoids the pitfalls of CRIL and DGR, ensuring a more accurate replication of the desired trajectory. The second row illustrates how each method generates trajectory data. CRIL generates the next state observation conditioned on the previous state observation. DGR, in contrast, does not attempt to generate a trajectory but generates individual state observations i.i.d. On the other hand, t-DGR generates state observations conditioned on the trajectory timestep.}
\label{fig:toy}
\end{figure*}

Drawing inspiration from the human brain, deep generative replay (DGR) addresses the catastrophic forgetting issue in decision-making tasks by using a generative model as the hippocampus to generate trajectories from past tasks and replay them to the learner which acts as the neocortex (Figure~\ref{fig:dgr}) \citep{shin2017continual}. The time-series nature of trajectories in decision-making tasks sets it apart from continual supervised learning, as each timestep of the trajectory requires sufficient replay. In supervised learning, the learner's performance is not significantly affected if it performs poorly on a small subset of the data. However, in decision-making tasks, poor performance on any part of the trajectory can severely impact the overall performance. Therefore, it is crucial to generate state-action pairs that accurately represent the distribution found in trajectories. Furthermore, the high-dimensional distribution space of trajectories makes it computationally infeasible to generate complete trajectories all at once.

Existing DGR methods adopt either the generation of individual state observations i.i.d. without considering the temporal nature of trajectories or autoregressive trajectory generation. Autoregressive approaches generate the next state(s) in a trajectory by modeling the conditional probability of the next state(s) given the previously generated state(s). However, autoregressive methods suffer from compounding errors in the generated trajectories. On the other hand, generating individual state observations i.i.d. leads to a higher sample complexity compared to generating entire trajectories, which becomes significant when replay time is limited (see Section~\ref{sec:rehearsal}).

To address the issues in current DGR methods, we propose a simple, scalable, and non-autoregressive trajectory-based DGR method. We define a generated trajectory as temporally coherent if the transitions from one state to the next appear realistic (refer to Section~\ref{sec:notation} for a formal definition). Given that imitation learning methods and many reinforcement learning methods are trained on single transition samples (e.g. state-action pairs), we do not require trajectories to exhibit temporal coherence. Instead, our focus is on ensuring an equal number of samples generated at each timestep of the trajectory to accurately represent the distribution found in trajectories. To achieve equal sample coverage at each timestep, we train our generator to produce state observations conditioned on the trajectory timestep, and then sample from the generator conditioned on each timestep of the trajectory. The intuition behind our method is illustrated in Figure~\ref{fig:toy}.

To evaluate the effectiveness of our proposed method, t-DGR, we conducted experiments on the Continual World benchmarks CW10 and CW20 \citep{wołczyk2021continual} using imitation learning. Our results indicate that t-DGR achieves state-of-the-art performance in terms of average success rate when compared to other top continual learning methods.

\section{Related Work}
This section provides an overview of existing continual learning methods, with a particular focus on pseudo-rehearsal methods.

\subsection{Continual Learning in the Real World}
As the field of continual learning continues to grow, there is an increasing emphasis on developing methods that can be effectively applied in real-world scenarios \citep{wang2024comprehensive, NEURIPS2019_e562cd9c, bang2021rainbow, hsu2019reevaluating, vandeven2019scenarios}. The concept of ``General Continual Learning" was introduced by \citet{buzzega2020dark} to address certain properties of the real world that are often overlooked or ignored by existing continual learning methods. Specifically, two important properties, bounded memory and blurry task boundaries, are emphasized in this work. Bounded memory refers to the requirement that the memory footprint of a continual learning method should remain bounded throughout the entire lifespan of the learning agent. This property is crucial to ensure practicality and efficiency in real-world scenarios. Additionally, blurry task boundaries highlight the challenge of training on tasks that are intertwined, without clear delineation of when one task ends and another begins. Many existing methods fail to account for this characteristic, which is common in real-world learning scenarios. While there are other significant properties associated with continual learning in the real world, this study focuses on the often-neglected aspects of bounded memory and blurry task boundaries. By addressing these properties, we aim to develop methods that are more robust and applicable in practical settings.

\subsection{Continual Learning Methods}
Continual learning methods for decision-making tasks can be categorized into three main categories.

\paragraph{Regularization}
Regularization methods in continual learning focus on incorporating constraints during model training to promote the retention of past knowledge. One simple approach is to include an $L_2$ penalty in the loss function. Elastic Weight Consolidation (EWC) builds upon this idea by assigning weights to parameters based on their importance for previous tasks using the Fisher information matrix \citep{Kirkpatrick_2017}. MAS measures the sensitivity of parameter changes on the model's output, prioritizing the retention of parameters with a larger effect \citep{aljundi2018memory}. VCL leverages variational inference to minimize the Kullback-Leibler divergence between the current and prior parameter distributions \citep{nguyen2018variational}. Progress and Compress learns new tasks using a separate model and subsequently distills this knowledge into the main model while safeguarding the previously acquired knowledge \citep{schwarz2018progress}. However, regularization methods may struggle with blurry task boundaries as they rely on knowledge of task endpoints to apply regularization techniques effectively. In our experiments, EWC was chosen as the representative regularization method based on its performance in the original Continual World experiments \citep{wołczyk2021continual}.

\paragraph{Architecture-based Methods}
Architecture-based methods aim to maintain distinct sets of parameters for each task, ensuring that future learning does not interfere with the knowledge acquired from previous tasks. Packnet \citep{mallya2018packnet}, UCL \citep{NEURIPS2019_2c3ddf4b}, and AGS-CL \citep{jung2021continual} all safeguard previous task information in a neural network by identifying important parameters and freeing up less important parameters for future learning. Identification of important parameters can be done through iterative pruning (Packnet), parameter uncertainty (UCL), and activation value (AGS-CL). However, a drawback of parameter isolation methods is that each task requires its own set of parameters, which may eventually exhaust the available parameters for new tasks and necessitate a dynamically expanding network without bounded memory \citep{yoon2018lifelong}. Additionally, parameter isolation methods require training on a single task at a time to prune and isolate parameters, preventing concurrent learning from multiple interwoven tasks. In our experiments, PackNet was selected as the representative architecture-based method based on its performance in the original Continual World experiments \citep{wołczyk2021continual}.

\begin{figure*}
\includegraphics[width=\linewidth]{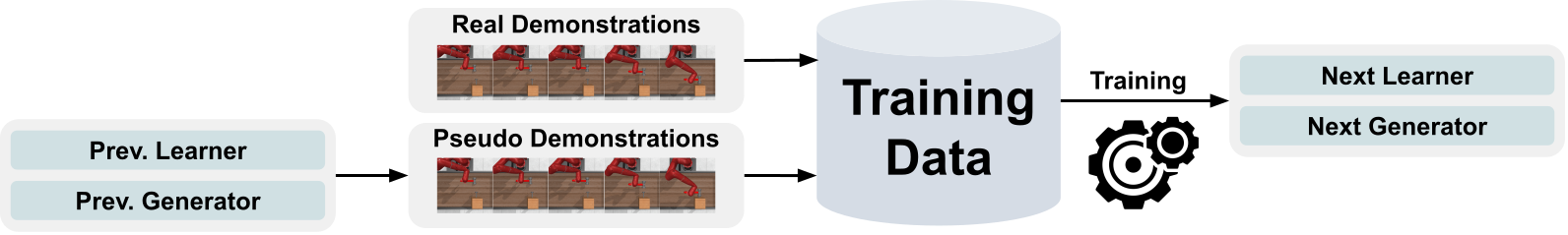}
\caption{The deep generative replay paradigm. The algorithm learns to generate trajectories from past tasks to augment real trajectories from the current task in order to mitigate catastrophic forgetting. Both the generator and policy model are updated with this augmented dataset.}
\label{fig:dgr}
\end{figure*}

\paragraph{Pseudo-rehearsal Methods}
\label{sec:rehearsal}
Pseudo-rehearsal methods mitigate the forgetting of previous tasks by generating synthetic samples from past tasks and replaying them to the learner. Deep generative replay (DGR) (Figure~\ref{fig:dgr}) utilizes a generative model, such as generative adversarial networks \citep{goodfellow2014generative}, variational autoencoders \citep{kingma2022autoencoding}, or diffusion models \citep{ho2020denoising, Sohl}, to generate the synthetic samples. Originally, deep generative replay was proposed to address continual supervised learning problems, where the generator only needed to generate single data point samples \citep{shin2017continual}. However, in decision-making tasks, expert demonstrations consist of trajectories (time-series) with a significantly higher-dimensional distribution space.

One existing DGR method generates individual state observations i.i.d. instead of entire trajectories. However, this approach leads to a higher sample complexity compared to generating entire trajectories. The sample complexity of generating enough individual state observations i.i.d. to cover every portion of the trajectory $m$ times can be described using the Double Dixie Cup problem \citep{doubleDixieCup}. For trajectories of length $n$, it takes an average of $\Theta(n \log n + m n \log\log n)$ i.i.d. samples to ensure at least $m$ samples for each timestep. In scenarios with limited replay time (small $m$) and long trajectories (large $n$) the sample complexity can be approximated as $\Theta(n \log n)$ using the Coupon Collector's problem \citep{couponCollector}.  The additional $\Theta(\log n)$ factor reduces the likelihood of achieving complete sample coverage of the trajectory when the number of replays or replay time is limited, especially considering the computationally expensive nature of current generative methods. Furthermore, there is a risk that the generator assigns different probabilities to each timestep of the trajectory, leading to a selective focus on certain timesteps rather than equal representation across the trajectory.

Another existing DGR method is autoregressive trajectory generation. In the existing autoregressive method, CRIL, a generator is used to generate samples of the initial state, and a dynamics model predicts the next state based on the current state and action \citep{gao2021cril}.  However, even with a dynamics model accuracy of 99\% and a 1\% probability of deviating from the desired trajectory, the probability of an autoregressively generated trajectory going off course is $1 - 0.99^n$, where $n$ denotes the trajectory length. With a trajectory length of $n = 200$ (as used in our experiments), the probability of an autoregressively generated trajectory going off course is $1 - 0.99^{200} = 0.87$. This example demonstrates how the issue of compounding error leads to a high probability of failure, even with a highly accurate dynamics model.

In our experiments, t-DGR is evaluated against all existing trajectory generation methods in pseudo-rehearsal approaches to assess how well t-DGR addresses the limitations of those methods.

\section{Background}
This section introduces notation and the formulation of the continual imitation learning problem that we use in this paper. Additionally, we provide a concise overview of diffusion probabilistic models used in our generative model implementation.

\subsection{Imitation Learning}
Imitation learning algorithms aim to learn a policy $\pi_\theta$ parameterized by $\theta$ by imitating a set of expert demonstrations $D = \{\tau_i\}_{i = 1 \ldots M}$. Each trajectory $\tau_i$ consists of a sequence of state-action pairs $\{(s_j, a_j)\}_{j = 1 \ldots |\tau_i|}$ where $|\tau_i|$ is the length of the trajectory. Each trajectory comes from a task $\mathcal{T}$ which is a Markov decision process that can be represented as a tuple $\langle S, A, T, \rho_0 \rangle$ with state space $S$, action space $A$, transition dynamics $T: S \times A \times S \to [0, 1]$, and initial state distribution $\rho_0$. Various algorithms exist for imitation learning, including behavioral cloning, GAIL \citep{ho2016generative}, and inverse reinforcement learning \citep{ng2000inverse}. In this work, we use behavioral cloning where the objective can be formulated as minimizing the loss function: 
\begin{equation} \label{eq:mseloss}
\mathcal{L}(\theta) = \mathbb{E}_{s,a \sim D}\bigg[\big\|\pi_\theta(s) - a\big\|^2_2 \bigg]
\end{equation}
where the state and action spaces are continuous.

\subsection{Continual Imitation Learning}
\label{sec:CL}
In the basic formulation most common in the field today, continual imitation learning involves sequentially solving multiple tasks $\mathcal{T}_1, \mathcal{T}_2, \ldots, \mathcal{T}_N$. When solving for task $\mathcal{T}_i$, the learner only gets data from task $\mathcal{T}_i$ and can not access data for any other task. In a more general scenario, certain tasks may have overlapping boundaries, allowing the learner to encounter training data from multiple tasks during certain phases of training. The learner receives a continuous stream of training data in the form of trajectories $\tau_1, \tau_2, \tau_3, \ldots$ from the environment, where each trajectory $\tau$ corresponds to one of the $N$ tasks. However, the learner can only access a limited contiguous portion of this stream at any given time. 

Let $s_i$ be the success rate of task $\mathcal{T}_i$ after training on all $N$ tasks. The continual imitation learning objective is defined as maximizing the average success rate over all tasks: 
\begin{equation} \label{eq:metric}
S = \frac{1}{N}\sum_{i = 1}^N s_i
\end{equation}
The primary issue that arises from the continual learning problem formulation is the problem of catastrophic forgetting where previously learned skills are forgotten when training on a new task. 

\subsection{Diffusion Probabilistic Models}
Diffusion probabilistic models \citep{ho2020denoising, Sohl} generate data through a learned reverse denoising diffusion process $p_\theta(x_{t - 1} \mid x_t)$. The forward diffusion process $q(x_t \mid x_{t - 1})$ gradually adds Gaussian noise to an input $x_0$ at each time step $t$, ultimately resulting in pure noise $x_T$ at $t = T$. The forward diffusion process is defined as:
\begin{equation}
    q(x_t \mid x_{t - 1}) = \mathcal{N}\left(x_t; \sqrt{1 - \beta_t}x_{t - 1}, \beta_t\mathbf{I}\right)
\end{equation}
where $0 < \beta_t < 1$ is defined by a known variance schedule. In our implementation, we adopted the cosine schedule proposed by \citet{nichol2021improved}. For a sufficiently large time horizon $T$ and a well-behaved variance schedule, $x_T$ approximates an isotropic Gaussian distribution. If we had the reverse diffusion process $p(x_{t - 1} \mid x_t)$, we could sample $x_T \sim \mathcal{N}(0, \mathbf{I})$ and obtain a sample from $q(x_0)$ by denoising $x_T$ with $p(x_{t - 1} \mid x_t)$. However, computing $p(x_{t - 1} \mid x_t)$ is intractable as it necessitates knowledge of the distribution of all possible $x_t$. Instead, we approximate $p(x_{t - 1} \mid x_t)$ using a neural network:
\begin{equation} \label{eq:ptheta}
    p_\theta(x_{t - 1} \mid x_t) = \mathcal{N}\left(x_{t - 1}; \mu_\theta(x_t, t), \Sigma(x_t, t)\right) 
\end{equation}
Since $q$ and $p_\theta$ can be viewed as a variational auto-encoder \citep{kingma2022autoencoding}, we can use the variational lower bound to minimize the negative log-likelihood of the reverse process. We can express $\mu_\theta(x_t, t)$ from Equation~\ref{eq:ptheta} as:
\begin{equation}
    \mu_\theta(x_t, t) = \frac{1}{\sqrt{\alpha_t}}\left(x_t - \frac{\beta_t}{\sqrt{1 - \overline{\alpha}_t}}\epsilon_\theta(x_t, t)\right)
\end{equation}
where $\alpha_t = 1 - \beta_t$ and $\overline{\alpha}_t = \prod_{s = 0}^t \alpha_s$. The training loss can then be defined as:
\begin{equation} \label{eq:diffLoss}
    \mathcal{L}(\theta) = \mathbb{E}_{x_0, t, \epsilon}\left[\lVert\epsilon - \epsilon_\theta(x_t, t)\rVert^2\right]
\end{equation}
Note that the timesteps $t$ in the diffusion process differ from the trajectory timesteps $t$. Henceforth, we will refer only to the trajectory timesteps $t$.

\subsection{Notation}
\label{sec:notation}
Deep generative replay involves training two models: a generator $G_\gamma$ parameterized by $\gamma$ and a learner $\pi_\theta$ parameterized by $\theta$. We define $G_\gamma^{(i)}$ as the generator trained on tasks $\mathcal{T}_1 \ldots \mathcal{T}_i$ and capable of generating data samples from tasks $\mathcal{T}_1 \ldots \mathcal{T}_i$. Similarly, $\pi_\theta^{(i)}$ represents the learner trained on tasks $\mathcal{T}_1 \ldots \mathcal{T}_i$ and able to solve tasks $\mathcal{T}_1 \ldots \mathcal{T}_i$.

A sequence of state observations $(s_1, s_2, \ldots, s_{n - 1}, s_n)$ is \textbf{temporally coherent} if $\forall 1 \leq i < n, \exists a \in A : T(s_i, a, s_{i + 1}) > \varepsilon$, where $0 < \varepsilon < 1$ is a small constant representing a threshold for negligible probabilities.

\section{Method}
Our proposed method, t-DGR, tackles the challenge of generating long trajectories by training a generator, denoted as $G_\gamma(j)$, which is conditioned on the trajectory timestep $j$ to generate state observations. Pseudocode for t-DGR is provided as Algorithm~\ref{alg:t-DGR}. The algorithm begins by initializing the task index, replay ratio, generator model, learner model, and learning rates (Line \ref{line:init}). The replay ratio, denoted as $0 \leq r < 1$, determines the percentage of training samples seen by the learner that are generated. Upon receiving training data from the environment, t-DGR calculates the number of trajectories to generate based on the replay ratio $r$ (Lines \ref{line:data}-\ref{line:n}). The variable $L$ (Line \ref{line:L}) represents the maximum length of trajectories observed so far.

To generate a trajectory $\tau$ of length $L$, t-DGR iterates over each timestep $1 \leq j \leq L$ (Line \ref{line:forL}). At each timestep, t-DGR generates the $j$-th state observation of the trajectory using the previous generator $G_\gamma^{(t - 1)}$ conditioned on timestep $j$ (Line \ref{line:gen}), and then labels it with an action using the previous policy $\pi_\theta^{(t - 1)}$ (Line \ref{line:label}). After generating all timesteps in the trajectory $\tau$, t-DGR adds it to the existing training dataset (Line \ref{line:add}). Note that the generated state observations within a trajectory do not have temporal coherence, as each state observation is generated independently of other timesteps. This approach is acceptable since our learner is trained on state-action pairs rather than full trajectories. However, unlike generating state observations i.i.d., our method ensures equal coverage of every timestep during the generative process, significantly reducing sample complexity.

Once t-DGR has augmented the training samples from the environment with our generated training samples, t-DGR employs backpropagation to update both the generator and learner using the augmented dataset (Lines \ref{line:beginBP}-\ref{line:endBP}). The t-DGR algorithm continues this process of generative replay throughout the agent's lifetime, which can be infinite (Line \ref{line:loop}). Although we perform the generative process of t-DGR at task boundaries for ease of understanding, no part of t-DGR is dependent on clear task boundaries.

\paragraph{Architecture}
We employ a U-net \citep{ronneberger2015u} trained on the loss specified in Equation~\ref{eq:diffLoss} to implement the generative diffusion model $G_\gamma$. Since we utilize proprioceptive observations in our experiments, $\pi_\theta$ is implemented with a multi-layer perceptron trained on the loss specified in Equation~\ref{eq:mseloss}.

\begin{algorithm}[t!]
\caption{Trajectory-based Deep Generative Replay (t-DGR)}
\label{alg:t-DGR}
\begin{algorithmic}[1]
\State Initialize task index $t = 0$, replay ratio $r$, generator $G^{(0)}_\gamma$, learner $\pi^{(0)}_\theta$, and learning rates $\lambda_\gamma, \lambda_\theta$. \label{line:init}
\While{new task available} \label{line:loop}
    \State $t \gets t + 1$
    \State Initialize dataset $D$ with trajectories from task $t$. \label{line:data}
    \State $n \gets \frac{r * |D|}{1 - r}$ \Comment{number of trajectories to generate} \label{line:n}
    \For{$i = 1$ to $n$}
        \State $L \gets$ maximum trajectory length \label{line:L}
        \State $\tau \gets \emptyset$ \Comment{initialize trajectory of length $L$}
        \For{$j = 1$ to $L$} \label{line:forL}
        \State $S \gets G_\gamma^{(t - 1)}(j)$ \Comment{generate states} \label{line:gen}
        \State $A \gets \pi_\theta^{(t - 1)}(S)$ \Comment{label with actions} \label{line:label}
        \State $\tau_j \gets (S, A)$ \Comment{add to trajectory}
        \EndFor
        \State $D \gets D \cup \tau$ \Comment{add generated trajectory to $D$} \label{line:add}
    \EndFor
    \State Update generator and learner using $D$ \label{line:beginBP}
    \State $\gamma^{(t)} \gets \gamma^{(t - 1)} - \lambda_\gamma \nabla_\gamma \mathcal{L}_{G^{(t - 1)}}(\gamma^{(t - 1)})$
    \State $\theta^{(t)} \gets \theta^{(t - 1)} - \lambda_\theta \nabla_\theta \mathcal{L}_{\pi^{(t - 1)}}(\theta^{(t - 1)})$ \label{line:endBP}
\EndWhile
\end{algorithmic}
\end{algorithm}

\section{Experiments}
In this section, we outline the experimental setup and performance metrics employed to compare t-DGR with representative methods, followed by an analysis of experimental results across different benchmarks and performance metrics.

\subsection{Experimental Setup}
We evaluate our method on the Continual World benchmarks CW10 and CW20 \citep{wołczyk2021continual}, along with our own variant of CW10 called BB10 that evaluates the ability of methods to handle blurry task boundaries. CW10 consists of a sequence of 10 Meta-World \citep{yu2021metaworld} tasks, where each task involves a Sawyer arm manipulating one or two objects in the Mujoco physics simulator. For computational efficiency, we provide the agents with proprioceptive observations. Notably, the observation and action spaces are continuous and remain consistent across all tasks. CW20 is an extension of CW10 with the tasks repeated twice. To our knowledge, Continual World is the only standard continual learning benchmark for decision-making tasks. BB10 gives data to the learner in 10 sequential buckets $B_1, \ldots, B_{10}$. Data from task $\mathcal{T}_i$ from CW10 is split evenly between buckets $B_{i - 1}$, $B_i$, and $B_{i + 1}$, except for the first and last task. Task $\mathcal{T}_1$ is evenly split between buckets $B_0$ and $B_1$, and task $\mathcal{T}_{10}$ is evenly split between buckets $B_9$ and $B_{10}$.

In order to bound memory usage, task conditioning should utilize fixed-size embeddings like natural language task embeddings, rather than maintaining a separate final neural network layer for each individual task. For simplicity, we condition the model on the task with a one-hot vector as a proxy for natural language prompt embeddings. For BB10, the model is still conditioned on the underlying task rather than the bucket. Additionally, we do not allow separate biases for each task, as originally done in EWC \citep{Kirkpatrick_2017}. Expert demonstrations for training are acquired by gathering 100 trajectories per task using hand-designed policies from Meta-World, with each trajectory limited to a maximum of 200 steps. Importantly, the learner model remains consistent across different methods and benchmark evaluations. Moreover, we maintain a consistent replay ratio of $r = 0.9$ across all pseudo-rehearsal methods. 

We estimated the success rate $S$ of a model by running each task 100 times. The metrics for each method were computed using 5 seeds to create a 90\% confidence interval. Further experimental details, such as hyperparameters, model architecture, random seeds, and computational resources, are included in the appendix. This standardization enables a fair and comprehensive comparison of our proposed approach with other existing methods.

\subsection{Metrics}
We evaluate our models using three metrics proposed by the Continual World benchmark \citep{wołczyk2021continual}, with the average success rate being the primary metric. Although the forward transfer and forgetting metrics are not well-defined in a setting with blurry task boundaries, they are informative within the context of Continual World benchmarks. As a reminder from Section~\ref{sec:CL}, let $N$ denote the number of tasks, and $s_i$ represent the success rate of the learner on task $\mathcal{T}_i$. Additionally, let $s_i(t)$ denote the success rate of the learner on task $\mathcal{T}_i$ after training on tasks $\mathcal{T}_1$ to $\mathcal{T}_t$.

\paragraph{Average Success Rate} The average success rate, as given by Equation~\ref{eq:metric}, serves as the primary evaluation metric for continual learning methods.

\paragraph{Average Forward Transfer} We introduce a slightly modified metric for forward transfer that applies to a broader range of continual learning problems beyond just continual reinforcement learning in the Continual World benchmark. Let $s_i^{\mathrm{ref}}$ represent the reference performance of a single-task experiment on task $\mathcal{T}_i$. The forward transfer metric $FT_i$ is computed as follows:
\begin{align*}
FT_i = \frac{D_i - D_i^{\mathrm{ref}}}{1 - D_i^{\mathrm{ref}}} && D_i = \frac{s_i(i) + s_i(i - 1)}{2} && D_i^{\mathrm{ref}} = \frac{s_i^{\mathrm{ref}}}{2}
\end{align*}
The expressions for $D_i$ and $D^{\mathrm{ref}}_i$ serve as approximations of the integral of task $\mathcal{T}_i$ performance with respect to the training duration for task $\mathcal{T}_i$. The average forward transfer $FT$ is then defined as the mean forward transfer over all tasks, calculated as $FT = \frac{1}{N}\sum_{i = 1}^N FT_i$.

\paragraph{Average Forgetting} We measure forgetting using the metric $F_i$, which represents the amount of forgetting for task $i$ after all training has concluded. $F_i$ is defined as the difference between the success rate on task $\mathcal{T}_i$ immediately after training and the success rate on task $\mathcal{T}_i$ at the end of training. \[F_i = s_i(i) - s_i(N)\] The average forgetting $F$ is then computed as the mean forgetting over all tasks, given by $F = \frac{1}{N}\sum_{i = 1}^N F_i$.

\subsection{Baselines}
We compare the following methods on the Continual World benchmark using average success rate as the primary evaluation metric. Representative methods were chosen based on their success in the original Continual World experiments, while DGR-based methods were selected to evaluate whether t-DGR addresses the limitations of existing pseudo-rehearsal methods.
\begin{itemize}
    \item \textbf{Finetune:} The policy is trained only on data from the current task.
    \item \textbf{Multitask:} The policy is trained on data from all tasks simultaneously.
    \item \textbf{oEWC~\citep{schwarz2018progress}:} A variation of EWC known as online Elastic Weight Consolidation (oEWC) bounds the memory of EWC by employing a single penalty term for the previous model instead of individual penalty terms for each task. This baseline is the representative regularization-based method.
    \item \textbf{PackNet~\citep{mallya2018packnet}:} This baseline is the representative parameter isolation method. Packnet safeguards previous task information in a neural network by iteratively pruning, freezing, and retraining parts of the network.
    \item \textbf{DGR~\citep{shin2017continual}:} This baseline is a deep generative replay method that only generates individual state observations i.i.d. and not entire trajectories.
    \item \textbf{CRIL~\citep{gao2021cril}:} This baseline is a deep generative replay method that trains a policy along with a start state generator and a dynamics model that predicts the next state given the current state and action. Trajectories are generated by using the dynamics model and policy to autoregressively generate next states from a start state.
    \item \textbf{t-DGR:} Our proposed method.
\end{itemize}
Due to the inability of oEWC and PackNet to handle blurry task boundaries, we made several adjustments for CW20 and BB10. Since PackNet cannot continue training parameters for a task once they have been fixed, we treated the second repetition of tasks in CW20 as distinct from the first iteration, resulting in PackNet being evaluated with $N = 20$, while the other methods were evaluated with $N = 10$. As for BB10 and its blurry task boundaries, the best approach we could adopt with oEWC and PackNet was to apply their regularization techniques at regular training intervals rather than strictly at task boundaries. During evaluation, all tasks were assessed using the last fixed set of parameters in the case of PackNet.

\subsection{Ablations}

To evaluate the effect of generator architecture on pseudo-rehearsal methods, we included an ablation where the diffusion models in pseudo-rehearsal methods are replaced with Wasserstein generative adversarial networks (GAN) with gradient penalty \citep{gulrajani2017improved}. To evaluate the effectiveness of diffusion models at generative replay, we evaluated the quality of generated samples for past tasks at all stages of continual learning. Since we utilize proprioceptive observations rather than RGB camera observations, there is no practical way to qualitatively evaluate the generated samples. The proprioceptive observations are 39-dimensional, comprising 3D positions and quaternions of objects, and the degree to which the robot gripper is open \citep{yu2021metaworld}. These raw numbers cannot be qualitatively evaluated like RGB images. Instead, we use the average diffusion loss as a quantitative proxy metric for generation quality. When the generator model is learning task $i$, we compute the average diffusion loss for data samples from tasks 1 to $i - 1$. We then compare this average diffusion loss to the average diffusion loss when the generator model first learned task $i$. The difference between the two provides an estimate of how much generative replay has degraded the generation quality of previous tasks.

\begin{table}[h]
    \captionsetup[subtable]{font=large}
  \centering
  \begin{subtable}{0.45\linewidth}
    \addtolength{\tabcolsep}{-1.5pt}
      \centering
      \caption{CW10}
      \begin{tabular}{l c c c c}
        \toprule
        Method & Success Rate $\uparrow$ & FT$\uparrow$ & Forgetting$\downarrow$ \\
        \midrule
        Finetune & 16.4 \fs{6.4} & -3.0 \fs{6.0} & 78.8 \fs{7.6} \\
        Multitask & 97.0 \fs{1.0} & N/A & N/A \\
        \midrule
        oEWC & 18.6 \fs{5.3} & -6.3 \fs{5.7} & 74.1 \fs{6.1} \\
        PackNet & 81.4 \fs{3.7} & -14.8 \fs{7.8} & \textbf{-0.1} \fs{1.2} \\
        DGR (gan) & 21.2 \fs{3.7} & -2.6 \fs{3.3} & 74.4 \fs{4.7} \\
        CRIL (gan) & 24.0 \fs{4.4} & 0.3 \fs{4.5} & 72.0 \fs{5.4} \\
        t-DGR (gan) & 17.4 \fs{4.4} & 0.1 \fs{3.1} & 79.8 \fs{4.2} \\
        DGR (diffusion) & 75.0 \fs{5.8} & -4.3 \fs{5.1} & 17.8 \fs{4.1} \\
        CRIL (diffusion) & 28.4 \fs{10.6} & -1.1 \fs{2.8} & 68.6 \fs{10.4} \\
        t-DGR (diffusion) & \textbf{81.9} \fs{3.3} & \textbf{-0.3} \fs{4.9} & 14.4 \fs{2.5} \\
        \bottomrule
      \end{tabular}
      
    \label{tbl:CW10}
  \end{subtable}
  \hfill
  \begin{subtable}{0.45\linewidth}
    \addtolength{\tabcolsep}{-1.5pt}
      \centering
      \caption{BB10}
      \begin{tabular}{l c}
        \toprule
        Method & Success Rate $\uparrow$ \\
        \midrule
        Finetune & 21.7 \fs{2.6} \\
        Multitask & 97.0 \fs{1.0} \\
        \midrule
        oEWC & 21.8 \fs{1.7} \\
        PackNet & 26.9 \fs{5.6} \\
        DGR (gan) & 35.6 \fs{4.3} \\
        CRIL (gan) & 37.5 \fs{4.7} \\
        t-DGR (gan) & 28.3 \fs{3.9} \\
        DGR (diffusion) & 75.3 \fs{4.4} \\
        CRIL (diffusion) & 53.5 \fs{5.5} \\
        t-DGR (diffusion) & \textbf{81.7}    \fs{4.0} \\
        \bottomrule
      \end{tabular}
      
    \label{tbl:BB10}
  \end{subtable}
  
    \vspace{1cm} 
    
  \begin{subtable}{0.45\linewidth}
    \addtolength{\tabcolsep}{-1.5pt}
      \centering
      \caption{CW20}
      \begin{tabular}{l c c c c}
        \toprule
        Method & Success Rate $\uparrow$ & FT$\uparrow$ & Forgetting$\downarrow$ \\
        \midrule
        Finetune & 14.2 \fs{4.0} & -0.5 \fs{3.0} & 82.2 \fs{5.6} \\
        Multitask & 97.0 \fs{1.0} & N/A & N/A \\
        \midrule
        oEWC & 19.4 \fs{5.3} & -2.8 \fs{4.1} & 75.2 \fs{7.5} \\
        PackNet & 74.1 \fs{4.1} & -20.4 \fs{3.4} & \textbf{-0.2} \fs{0.9} \\
        DGR (gan) & 19.0 \fs{2.1} & 0.8 \fs{2.3} & 78.6 \fs{2.3} \\
        CRIL (gan) & 26.7 \fs{6.5} & -0.6 \fs{0.7} & 72.1 \fs{6.2} \\
        t-DGR (gan) & 20.6 \fs{6.1} & 0.7 \fs{3.4} & 76.3 \fs{5.3} \\
        DGR (diffusion) & 74.1 \fs{4.1} & 18.9 \fs{2.9} & 23.3 \fs{3.3} \\
        CRIL (diffusion) & 50.8 \fs{4.4} & 4.4 \fs{4.9} & 46.1 \fs{5.4} \\
        t-DGR (diffusion) & \textbf{83.9} \fs{3.0} & \textbf{30.6} \fs{4.5} & 14.6 \fs{2.9} \\
        \bottomrule
      \end{tabular}
      
        \label{tbl:CW20}
  \end{subtable}
  \hfill
  \begin{subtable}{0.45\linewidth}
    \addtolength{\tabcolsep}{-1.5pt}
      \centering
      \caption{Replay Ratio}
      \begin{tabular}{c|c c c}
        \toprule
        Ratio & t-DGR & DGR \\
        \midrule
        0.5 & \textbf{63.2} \fs{2.6} & 52.8 \fs{2.9} \\
        0.6 & \textbf{66.3} \fs{4.4} & 56.9 \fs{4.5} \\
        0.7 & \textbf{70.8} \fs{4.1} & 62.5 \fs{3.6} \\
        0.8 & \textbf{75.0} \fs{6.9} & 69.2 \fs{4.9} \\
        0.9 & \textbf{81.9} \fs{3.3} & 75.0 \fs{5.8} \\
        \bottomrule
      \end{tabular}
      
        \label{tbl:ratio}
  \end{subtable}

  \vspace{1cm} 
  
  \caption{Tables (a), (b), and (c) present the results for Continual World 10, Blurry Boundaries 10, and Continual World 20, respectively. The tables display the average success rate, forward transfer, and forgetting (if applicable) with 90\% confidence intervals using 5 random seeds. An up arrow indicates that higher values are better and a down arrow indicates that smaller values are better. Table (d) compares the impact of replay amount on the average success rate of t-DGR and DGR on CW10 with 90\% confidence intervals obtained using 5 random seeds. The best results are highlighted in bold.}
  \label{tab:all_tables}
\end{table}

\subsection{Discussion}
t-DGR emerges as the leading method, demonstrating the highest success rate on CW10 (Table~\ref{tbl:CW10}), CW20 (Table~\ref{tbl:CW20}), and BB10 (Table~\ref{tbl:BB10}). Notably, PackNet's performance on the second iteration of tasks in CW20 diminishes, highlighting its limited capacity for continually accommodating new tasks. This limitation underscores the fact that PackNet falls short of being a true lifelong learner, as it necessitates prior knowledge of the task count for appropriate parameter capacity allocation. On the contrary, pseudo-rehearsal methods, such as t-DGR, exhibit improved performance with the second iteration of tasks in CW20 due to an increased replay time. These findings emphasize the ability of DGR methods to effectively leverage past knowledge, as evidenced by their superior forward transfer in both CW10 and CW20.

BB10 (Table~\ref{tbl:BB10}) demonstrates that pseudo-rehearsal methods are mostly unaffected by blurry task boundaries, whereas PackNet's success rate experiences a significant drop-off. This discrepancy arises from the fact that PackNet's regularization technique does not work effectively with less clearly defined task boundaries.

In our experiments across CW10, CW20, and BB10, we observed that diffusion models outperform GANs as the generator for pseudo-rehearsal methods. We hypothesize that the distributional shifts in continual learning exacerbate instability issues with GAN training \citep{salimans2016improved, brock2018large, miyato2018spectral}. The motivation behind CRIL as stated in the paper \citep{gao2021cril} is to alleviate the burden of trajectory generation from the generator by transferring some of the generation complexity to a dynamics model. Our findings support this motivation when working with less capable generators \citep{dhariwal2021diffusion}, as reducing the generator's burden consistently enhances performance of GAN-based pseudo-rehearsal methods across all benchmarks. Among pseudo-rehearsal techniques, CRIL places the least demand on its generator, which only needs to produce the initial frame of a trajectory. In contrast, DGR requires generation of every frame, and t-DGR, the most demanding, requires generation of every frame along with the handling of trajectory timestep conditioning. Results from our GAN-based pseudo-rehearsal experiments on CW10, CW20, and BB10 indicate that CRIL outperforms DGR, which in turn outperforms t-DGR. However, experiments using a more capable diffusion generator \citep{dhariwal2021diffusion} reveal a reversal in performance ranking among these methods, suggesting that more capable generators diminish the need to offload generation complexity from the main generator.

The diminishing performance gap between DGR and t-DGR as the replay ratio increases in Table~\ref{tbl:ratio} indicates that a higher replay ratio reduces the likelihood of any portion of the trajectory being insufficiently covered when sampling individual state observations i.i.d., thereby contributing to improved performance. This trend supports the theoretical sample complexity of DGR derived in Section~\ref{sec:rehearsal}, as $\Theta(n \log n + m n \log\log n)$ closely approximates the sample complexity of t-DGR, $\Theta(mn)$, when the replay amount $m \to \infty$. However, while DGR can achieve comparable performance to t-DGR with a high replay ratio, the availability of extensive replay time is often limited in many real-world applications.

\begin{figure*}[h]
\includegraphics[width=\linewidth]{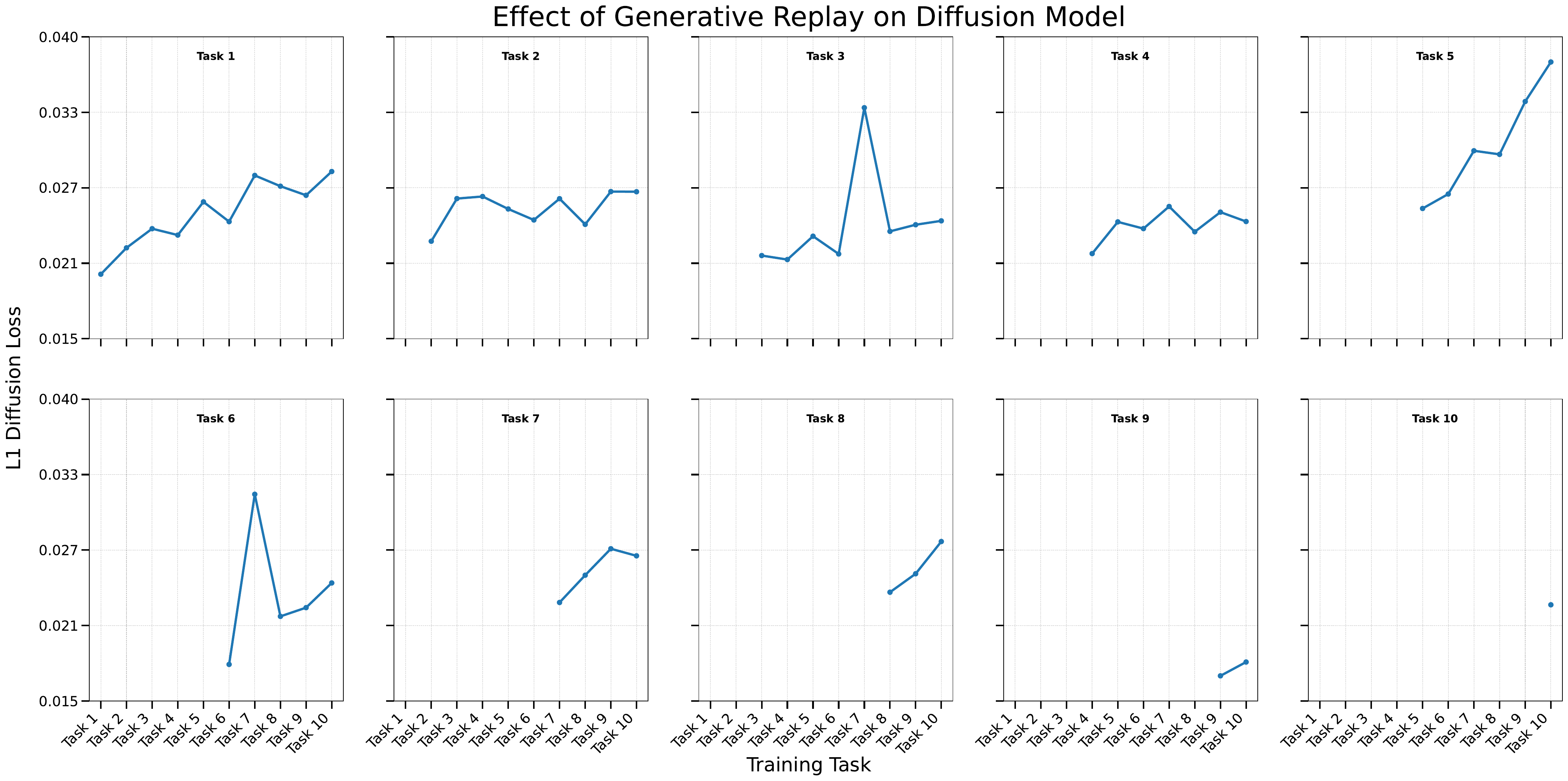}
\caption{This table illustrates the ability of the diffusion model in t-DGR to generate past data as it continues to learn additional tasks in CW10 through generative replay. The line plot for task $i$ plots the average diffusion loss of the diffusion model in future tasks on task $i$ data. The loss is an L1 version of the diffusion training loss in Equation~\ref{eq:diffLoss}.}
\label{fig:curve}
\end{figure*}

Recent studies have indicated that applying generative replay to diffusion models for image data leads to a collapse of generation quality due to compounding errors in the generated synthetic data \citep{Zajac2023ExploringCL, masip2023continual, smith2023continual}. However, our experiments reveal that when generating lower dimensional proprioceptive data, the diffusion model is able to maintain generation quality. Although Figure~\ref{fig:curve} shows that generative replay degrades generation quality of past tasks as the diffusion model learns new tasks, the success rates in Table~\ref{tab:all_tables} suggest that this degradation in generation quality is not severe enough to impact the learner's performance. Notably, generation quality for past tasks appears to deteriorate in tasks 5 and 7 but improves in the subsequent task, suggesting that the error compounding in diffusion models during generative replay might not be as severe as previously assumed, particularly for lower-dimensional data.

Overall, t-DGR exhibits promising results, outperforming other methods in terms of success rate in all evaluations. Notably, t-DGR achieves a significant improvement over existing pseudo-rehearsal methods on CW20 using a Welch t-test with a significance level of $\text{p-value} = 0.005$. Its ability to handle blurry task boundaries, leverage past knowledge, and make the most of replay opportunities position it as a state-of-the-art method for continual lifelong learning in decision-making.

\section{Conclusion}
In conclusion, we have introduced t-DGR, a novel method for continual learning in decision-making tasks, which has demonstrated state-of-the-art performance on the Continual World benchmarks. Our approach stands out due to its simplicity, scalability, and non-autoregressive nature, positioning it as a solid foundation for future research in this domain.

Importantly, t-DGR takes into account essential properties of the real world, including bounded memory and blurry task boundaries. These considerations ensure that our method remains applicable and effective in real-world scenarios, enabling its potential integration into practical applications.

Looking ahead, one potential avenue for future research is the refinement of the replay mechanism employed in t-DGR. Rather than assigning equal weight to all past trajectories, a more selective approach could be explored. By prioritizing certain memories over others and strategically determining when to replay memories to the learner, akin to human learning processes, we could potentially enhance the performance and adaptability of our method.

\section*{Acknowledgements}

\thanks{This work has taken place in the Learning Agents Research
Group (LARG) at the Artificial Intelligence Laboratory, The University
of Texas at Austin.  LARG research is supported in part by the
National Science Foundation (FAIN-2019844, NRT-2125858), the Office of
Naval Research (N00014-18-2243), Army Research Office (E2061621),
Bosch, Lockheed Martin, and Good Systems, a research grand challenge
at the University of Texas at Austin.  The views and conclusions
contained in this document are those of the authors alone.  Peter
Stone serves as the Executive Director of Sony AI America and receives
financial compensation for this work.  The terms of this arrangement
have been reviewed and approved by the University of Texas at Austin
in accordance with its policy on objectivity in research.}

\bibliography{references}
\bibliographystyle{collas2024_conference}

\newpage
\appendix
\section*{\centering Appendix}

\section{Hyperparameters}

\subsection{Finetune}

\begin{table}[H]
    \centering
    \label{tab:param:finetune}
    \begin{tabular}{|lcc|}
        \hline
        \textbf{Hyperparameter} & \textbf{Value} & \textbf{Brief Description} \\
        \hline
        batch size & 32 & number of samples in each training iteration \\
        epochs & 250 & number of times the entire dataset is passed through per task \\
        learning rate & $10^{-4}$ & learning rate for gradient descent \\
        optimization algorithm & Adam & optimization algorithm used \\
        $\beta_1$ & 0.9 & exponential decay rate for first moment estimates in Adam \\
        $\beta_2$ & 0.999 & exponential decay rate for second moment estimates in Adam \\
        epsilon & $10^{-8}$ & small constant for numerical stability \\
        weight decay & 0 & weight regularization \\
        \hline
    \end{tabular}
\end{table}

\subsection{Multitask}

\begin{table}[H]
    \centering
    \label{tab:param:multitask}
    \begin{tabular}{|lcc|}
        \hline
        \textbf{Hyperparameter} & \textbf{Value} & \textbf{Brief Description} \\
        \hline
        batch size & 32 & number of samples in each training iteration \\
        epochs & 500 & number of times the entire dataset of all tasks is passed through \\
        learning rate & $10^{-4}$ & learning rate for gradient descent \\
        optimization algorithm & Adam & optimization algorithm used \\
        $\beta_1$ & 0.9 & exponential decay rate for first moment estimates in Adam \\
        $\beta_2$ & 0.999 & exponential decay rate for second moment estimates in Adam \\
        epsilon & $10^{-8}$ & small constant for numerical stability \\
        weight decay & 0 & weight regularization \\
        \hline
    \end{tabular}
\end{table}

\subsection{oEWC}

\begin{table}[H]
    \centering
    \label{tab:param:owec}
    \begin{tabular}{|lcc|}
        \hline
        \textbf{Hyperparameter} & \textbf{Value} & \textbf{Brief Description} \\
        \hline
        batch size & 32 & number of samples in each training iteration \\
        epochs & 250 & number of times the entire dataset of all tasks is passed through \\
        learning rate & $10^{-4}$ & learning rate for gradient descent \\
        Fisher multiplier & $10^2$ & the Fisher is scaled by this number to form the EWC penalty \\
        optimization algorithm & Adam & optimization algorithm used \\
        $\beta_1$ & 0.9 & exponential decay rate for first moment estimates in Adam \\
        $\beta_2$ & 0.999 & exponential decay rate for second moment estimates in Adam \\
        epsilon & $10^{-8}$ & small constant for numerical stability \\
        weight decay & 0 & weight regularization \\
        \hline
    \end{tabular}
\end{table}

The Fisher multiplier hyperparameter was tuned with the values: \(10^{-2}, 10^{-1}, 10^{0}, 10^1, 10^2, 10^3, 10^4, 10^5, 10^6\). We selected the value $10^2$ based on the success rate metric given by Equation~2. 

\subsection{PackNet}

\begin{table}[H]
    \centering
    \label{tab:param:packnet}
    \begin{tabular}{|lcc|}
        \hline
        \textbf{Hyperparameter} & \textbf{Value} & \textbf{Brief Description} \\
        \hline
        batch size & 32 & number of samples in each training iteration \\
        epochs & 250 & number of times the entire dataset of all tasks is passed through \\
        retrain epochs & 125 & number of training epochs after pruning \\
        learning rate & $10^{-4}$ & learning rate for gradient descent \\
        prune percent & 0.75 & percent of free parameters pruned for future tasks \\
        optimization algorithm & Adam & optimization algorithm used \\
        $\beta_1$ & 0.9 & exponential decay rate for first moment estimates in Adam \\
        $\beta_2$ & 0.999 & exponential decay rate for second moment estimates in Adam \\
        epsilon & $10^{-8}$ & small constant for numerical stability \\
        weight decay & 0 & weight regularization \\
        \hline
    \end{tabular}
\end{table}

The retrain epochs and prune percent hyperparameters were chosen following the approach in the original PackNet paper. After training the first task, bias layers are frozen.

\subsection{DGR}

\begin{table}[H]
    \centering
    \label{tab:param:dgr}
    \begin{tabular}{|lcc|}
        \hline
        \textbf{Hyperparameter} & \textbf{Value} & \textbf{Brief Description} \\
        \hline
        batch size & 32 & number of samples in each training iteration \\
        epochs & 250 & number of times the entire dataset of all tasks is passed through \\
        learning rate & $10^{-4}$ & learning rate for gradient descent \\
        diffusion training steps & $10^4$ & number of training steps for the diffusion model per task \\ 
        diffusion warmup steps & $5 * 10^4$ & number of extra training steps for the diffusion model on the first task \\
        diffusion timesteps & $10^3$ & number of timesteps in the diffusion process \\ 
        replay ratio & 0.9 & percentage of training examples that are generated \\
        optimization algorithm & Adam & optimization algorithm used \\
        $\beta_1$ & 0.9 & exponential decay rate for first moment estimates in Adam \\
        $\beta_2$ & 0.999 & exponential decay rate for second moment estimates in Adam \\
        epsilon & $10^{-8}$ & small constant for numerical stability \\
        weight decay & 0 & weight regularization \\
        \hline
    \end{tabular}
\end{table}

\subsection{CRIL}

\begin{table}[H]
    \centering
    \label{tab:param:cril}
    \begin{tabular}{|lcc|}
        \hline
        \textbf{Hyperparameter} & \textbf{Value} & \textbf{Brief Description} \\
        \hline
        batch size & 32 & number of samples in each training iteration \\
        epochs & 300 & number of times the entire dataset of all tasks is passed through \\
        learning rate & $10^{-4}$ & learning rate for gradient descent \\
        diffusion training steps & $10^4$ & number of training steps for the diffusion model per task \\ 
        diffusion warmup steps & $5 * 10^4$ & number of extra training steps for the diffusion model on the first task \\
        diffusion timesteps & $10^3$ & number of timesteps in the diffusion process \\ 
        replay ratio & 0.9 & percentage of training examples that are generated \\
        optimization algorithm & Adam & optimization algorithm used \\
        $\beta_1$ & 0.9 & exponential decay rate for first moment estimates in Adam \\
        $\beta_2$ & 0.999 & exponential decay rate for second moment estimates in Adam \\
        epsilon & $10^{-8}$ & small constant for numerical stability \\
        weight decay & 0 & weight regularization \\
        \hline
    \end{tabular}
\end{table}

\subsection{t-DGR}

\begin{table}[H]
    \centering
    \label{tab:param:tdgr}
    \begin{tabular}{|lcc|}
        \hline
        \textbf{Hyperparameter} & \textbf{Value} & \textbf{Brief Description} \\
        \hline
        batch size & 32 & number of samples in each training iteration \\
        epochs & 300 & number of times the entire dataset of all tasks is passed through \\
        learning rate & $10^{-4}$ & learning rate for gradient descent \\
        diffusion training steps & $10^4$ & number of training steps for the diffusion model per task \\ 
        diffusion warmup steps & $5 * 10^4$ & number of extra training steps for the diffusion model on the first task \\
        diffusion timesteps & $10^3$ & number of timesteps in the diffusion process \\ 
        replay ratio & 0.9 & percentage of training examples that are generated \\
        optimization algorithm & Adam & optimization algorithm used \\
        $\beta_1$ & 0.9 & exponential decay rate for first moment estimates in Adam \\
        $\beta_2$ & 0.999 & exponential decay rate for second moment estimates in Adam \\
        epsilon & $10^{-8}$ & small constant for numerical stability \\
        weight decay & 0 & weight regularization \\
        \hline
    \end{tabular}
\end{table}

\section{Model Architecture}

\subsection{Multi-layer Perceptron}

\begin{table}[H]
    \centering
    \begin{tabular}{|l|l|l|}
        \hline
        \textbf{Layer (type)} & \textbf{Output Shape} & \textbf{Param \#} \\
        \hline
        Linear-1 & [32, 512] & 25,600 \\
        Linear-2 & [32, 512] & 262,656 \\
        Linear-3 & [32, 512] & 262,656 \\
        Linear-4 & [32, 512] & 262,656 \\
        Linear-5 & [32, 4]   & 2,052 \\
        \hline
        \multicolumn{2}{|r|}{Total params:} & 815,620 \\
        \multicolumn{2}{|r|}{Trainable params:} & 815,620 \\
        \multicolumn{2}{|r|}{Non-trainable params:} & 0 \\
        \hline
    \end{tabular}
    \caption{Multi-layer perceptron architecture of the learner shared by all methods, featuring five linear layers with ReLU activations applied after each layer except the final one.}
    \label{tab:model_architecture}
\end{table}

To condition the MLP learner on the task, the input vector is concatenated with a one-hot vector representing the task before being passed into the MLP.

\subsection{Diffusion U-net}

The U-net consists of 4 downsampling and 4 upsampling layers. To condition the generator U-net on the trajectory timestep, we use a sinusoidal positional encoder from Transformers to create an sinusoidal positional embedding that gets passed through 3 linear layers and then added to the convolutional layer output at each level of the U-net.

\subsection{Generative Adversarial Network}

The generator is implemented with an MLP consisting of 7 linear layers with batch normalization and Leaky ReLU activations applied after each layer except the final one. 

The discriminator is implemented with an MLP consisting of 3 linear layers with Leaky ReLU activations applied after each layer except the final one.

Sinusoidoal positional encodings are fed as input to a 2-layer MLP to condition both the generator and discriminator on the trajectory timestep for t-DGR.

\section{Experiment Details}

We utilized the following random seeds for the experiments: 1, 2, 3, 4, 5. All experiments were conducted on Nvidia A100 GPUs with 80 GB of memory. The computational node consisted of an Intel Xeon Gold 6342 2.80GHz CPU with 500 GB of memory. For our longest benchmark, CW20, the runtimes were as follows: DGR and t-DGR took 3 days, CRIL took 16 hours, finetune and oEWC took 6 hours, and PackNet took 8 hours.

\end{document}